\pdfoutput=1

\documentclass[11pt]{article}

\usepackage[final]{acl}
\usepackage{todonotes}
\usepackage{xspace}
\usepackage{booktabs}
\usepackage{subcaption}
\usepackage{tabularray}
\usepackage{multirow}
\usepackage{hhline}
\usepackage{xcolor, colortbl}
\usepackage{listings}

\usepackage{amsmath}
\usepackage{amsfonts} 
\usepackage{algorithm}
\usepackage{algorithmic}
\usepackage{wrapfig}
\usepackage{float}
\usepackage{pifont}
\usepackage{newunicodechar}
\usepackage{enumitem}

\definecolor{Gray}{gray}{0.92}

\newcommand{\rparagraph}[1]{\vspace{1.2mm}\noindent\textbf{#1.}}

\newcommand{\sparagraph}[1]{\vspace{0.0mm}\noindent\textbf{#1.}}

\newcommand{\ours}{\textsc{ZEPO}\xspace} %

\newcommand{\oursfullx}{\textbf{Z}ero-shot \textbf{E}valuation-oriented \textbf{P}rompt \textbf{O}ptimization\xspace}

\makeatletter
\def\hlinewd#1{%
\noalign{\ifnum0=`}\fi\hrule \@height #1 %
\futurelet\reserved@a\@xhline}
\makeatother

\definecolor{racing-green}{RGB}{0, 185, 0}
\definecolor{awesome-red}{RGB}{255, 94, 94}
\definecolor{g-red}{RGB}{213, 66, 56}
\definecolor{g-green}{RGB}{49, 149, 79}
\definecolor{g-purple}{RGB}{181, 116, 157}
\definecolor{g-brown}{RGB}{232, 140, 79}

\definecolor{wan-brown}{RGB}{255, 245, 235}

\definecolor{wan-purple}{RGB}{229, 229, 255}

\definecolor{g-grey}{RGB}{153, 153, 153}

\newcommand{\yellowbg}[1]{%
  \begingroup\setlength{\fboxsep}{0pt}%
  \colorbox{wan-brown}{#1}%
  \endgroup
}

\newcommand{\bluebg}[1]{%
  \begingroup\setlength{\fboxsep}{0pt}%
  \colorbox{wan-purple}{#1}%
  \endgroup
}

\usepackage{times}
\usepackage{latexsym}

\usepackage[T1]{fontenc}

\usepackage[utf8]{inputenc}

\usepackage{microtype}

\usepackage{inconsolata}

\usepackage{graphicx}

\title{Fairer Preferences Elicit Improved Human-Aligned \\Large Language Model Judgments}

\author{Han Zhou\textsuperscript{1}
\quad
Xingchen Wan\textsuperscript{2}\thanks{Now at Google. Code is available at \url{https://github.com/cambridgeltl/zepo}.}
\quad
Yinhong Liu\textsuperscript{1}
\quad 
Nigel Collier\textsuperscript{1}
\quad \\
\textbf{Ivan Vuli{\'c}\textsuperscript{1}}
\quad
\textbf{Anna Korhonen\textsuperscript{1}}
 \\
  \textsuperscript{1}Language Technology Lab, University of Cambridge \\
  \textsuperscript{2}Machine Learning Research Group, University of Oxford \\
  \texttt{\{hz416, yl535, nhc30, iv250, alk23\}@cam.ac.uk}}

\begin{document}
\maketitle
\begin{abstract}
Large language models (LLMs) have shown promising abilities as cost-effective and reference-free evaluators for assessing language generation quality. In particular, pairwise LLM evaluators, which compare two generated texts and determine the preferred one, have been employed in a wide range of applications. However, LLMs exhibit preference biases and worrying sensitivity to prompt designs. In this work, we first reveal that the predictive preference of LLMs can be highly brittle and skewed, even with semantically equivalent instructions. We find that \textit{fairer} predictive preferences from LLMs consistently lead to judgments that are better aligned with humans. Motivated by this phenomenon, we propose an automatic \oursfullx framework, \ours, which aims to produce fairer preference decisions and improve the alignment of LLM evaluators with human judgments. To this end, we propose a zero-shot learning objective based on the preference decision \textit{fairness}. \ours demonstrates substantial performance improvements over state-of-the-art LLM evaluators, \textit{without} requiring labeled data, on representative meta-evaluation benchmarks. Our findings underscore the critical correlation between preference fairness and human alignment, positioning \ours as an efficient prompt optimizer for bridging the gap between LLM evaluators and human judgments.

\end{abstract}
\section{Introduction}
Large language models (LLMs) \citep{brown2020language, openai2023gpt4, team2023gemini, anil2023palm} have become the standard machinery for evaluating the quality of natural language generation over various aspects, such as coherence, fluency, and truthfulness, in a reference-free manner \citep{chen-etal-2023-exploring-use, zeng2024evaluating, zheng2024judging}. 
Owing to the remarkable in-context learning capabilities of LLMs \citep{brown2020language}, prompting techniques further enable versatile use of LLM evaluators with user-defined evaluation criteria, where pairwise-preference-based evaluators have so far demonstrated superior human alignment to direct scoring~\citep{liusie-etal-2024-llm, liu2024aligning}.  

\begin{figure}
    \centering
    \includegraphics[width=0.95\linewidth]{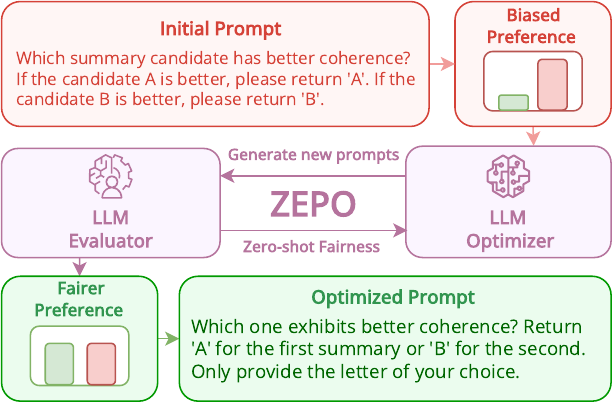}
    \vspace{-1mm}
    \caption{Illustration of the \textcolor{g-purple}{\textbf{\ours}} pipeline. Given a manual prompt, the distribution of LLM preferences can be \textcolor{g-red}{\textit{biased}} towards a certain class. \ours optimizes the prompt on a \textcolor{g-purple}{zero-shot fairness} learning objective until the \textcolor{g-green}{\textit{balance}} is achieved in the distribution.}
    \label{fig:main}
    \vspace{-2mm}
\end{figure}

However, LLMs have been known to exhibit preference bias \citep{wang2023large}, a priori propensity to predict certain classes over others unfairly, and display strong sensitivity to the actual prompts describing evaluation criteria \citep{zhou-etal-2023-survival, sclar2024quantifying}. The preference bias is argued to be largely due to various factors that result in a label distribution shift, such as position bias \citep{zheng2024judging}, verbosity bias \citep{saito2023verbosity}, and contextual bias \citep{zhou2024batch}, where LLMs unfairly favor later and longer answers, or even follow repetitive answers in their demonstrations. We are thus motivated to explore the impact of preference biases on human alignment in the novel context of LLM evaluators. 
We start by conducting a systematic study examining the sensitivity of LLM evaluators to the provided instructions. By paraphrasing from a set of instructions, we find that the pairwise preference of LLMs largely varies even with semantically equivalent instructions, and different instructions exert different degrees of preference biases. Noticeably, we observe that fairer preferences consistently lead to better human-aligned judgments. 
Motivated by this empirical finding, we then propose an automatic \textit{\oursfullx} (\ours) framework for steering LLM evaluators towards better agreements with humans; see Fig.~\ref{fig:main}. We design a new zero-shot fairness objective function by measuring the absolute difference between a uniform prior distribution and the model preference distribution. \ours, without any labeled data, shows substantial performance gains over state-of-the-art LLM evaluators with manually designed instructions on meta-evaluation benchmarks.

In sum, we provide the following contributions.
    \textbf{1)} We present a systematic analysis that reveals the strong sensitivity of LLM evaluators to instructions. Importantly, we find that \textit{fairer preferences elicit better human-aligned LLM judgments}.
    \textbf{2)} We introduce a \oursfullx framework (\ours) for automatically optimizing LLM evaluators toward better human agreements without any labeled data. 
    \textbf{3)} We demonstrate that \ours efficiently discovers the fairest instruction for LLM evaluators, delivering substantial gains in evaluation over representative tasks.

\begin{figure}[!t]
    \centering
    \begin{subfigure}[b]{0.445\linewidth}
        \centering
        \includegraphics[width=\textwidth]{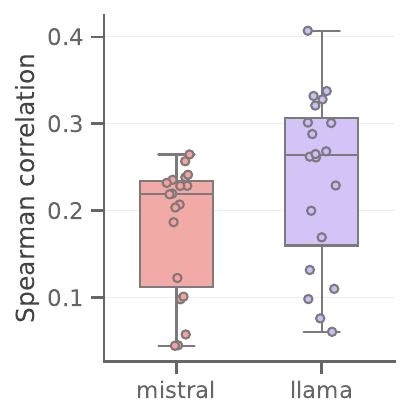}
        \caption*{}
    \end{subfigure}%
    \begin{subfigure}[b]{0.555\linewidth}
        \centering
        \includegraphics[width=\textwidth]{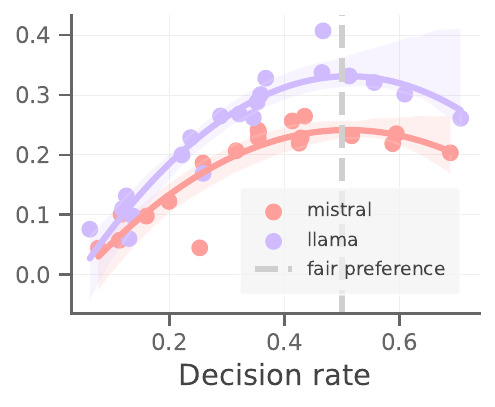}
        \caption*{}
    \end{subfigure}
    \vspace{-14mm}
    \caption{\textit{LLM evaluators show strong sensitivity to instructions and \textcolor{g-grey}{\textbf{fairer preference}} leads to better human-aligned LLM judgments.} Sensitivity and evaluation performance studies on preference fairness.}
    \label{fig:sensitivity}
\end{figure}

\section{Related Work}
\sparagraph{LLMs as Evaluators}
LLMs have been widely used to evaluate natural language generation tasks 
\citep{zhong-etal-2022-towards, chiang-lee-2023-large}, enabling automatic and reference-free evaluations \citep{liu-etal-2023-g, fu2023gptscore, chen-etal-2023-exploring-use,dong2024can}. Recent studies show that LLM evaluators can serve as effective pairwise text rankers
\citep{qin2023large}, where pairwise comparisons lead to better human-aligned judgments than Likert-score evaluations \citep{liusie-etal-2024-llm, liu2024aligning}. Yet, there is still a prominent gap between LLM evaluators and human agreement~\citep{shen-etal-2023-large}. LLM evaluators are yet sensitive to exemplars \citep{wang2023large} and exhibit unfair predictions due to position bias, verbosity bias, and self-preferences~\citep{zheng2024judging, pezeshkpour2023large, panickssery2024llm,liu2024measuring}. Calibration methods have been proposed to alleviate biases \citep{li2023split,li2024can, zhou2024batch}, but are yet insufficient for addressing all aforementioned biases.
In this work, we show that instructions exert large impacts on LLM evaluators, and searching for instructions with fairer preferences is a necessary and critical component in LLM-based evaluators.

\rparagraph{Automatic Prompt Optimization}
Unlike soft prompt tuning that requires `white box' access to model parameters~\citep{lester-etal-2021-power, zhou2024autopeft}, hard prompt tuning directly searches for discrete prompts that are portable and `black box' \citep{deng-etal-2022-rlprompt, zhou-etal-2023-survival}. Recent prompt optimization work further leverages LLMs as optimizers to generate more human interpretable prompts \citep{zhou2023large, yang2024large}. Much effort has been devoted to more advanced search algorithms \citep{pryzant-etal-2023-automatic, guo2024connecting, khattab2024dspy, wan2024teach, liu-etal-2024-calibrating-llm} but they heavily rely on labeled data. Instead, \textit{zero-shot} prompt optimization is a rather underexplored research area, and previous work is mostly limited to entropy-based exemplar selection \citep{lu-etal-2022-fantastically} or relies on model-synthesized data~\citep{chen-etal-2023-self}. We explore the extreme, zero-shot learning setup and leverage LLM's self-predictive distribution to optimize toward fairer preferences. As we will show, our fairness objective shows the best correlation and outweighs other zero-shot metrics for LLM evaluators in Fig.~\ref{fig:metrics}.

\section{Fairer Preferences Elicit Improved Human-Aligned Judgments}

\sparagraph{Prompt Sensitivity and Bias}
We start by analyzing the sensitivity of LLM evaluators to variations in instructions. 
Formally, given some source text and corresponding response candidates as an input query $x_i$, we have the predicted label $y_i$ as the model preference. Evaluation instruction $I$ is formulated with the input query $x_i$ in a prompt template to form a complete context $C(x_i, I) = \texttt{Template}(x_i,I)$ for evaluation. LLM evaluators then make predictions by $y_i = \operatorname*{arg\,max}_{y\in \mathcal{Y}}p(y|C_i)$, where the verbalizer $\mathcal{Y}$ defines the set of preferences (i.e., A or B for pairwise preferences). To inspect prompt sensitivity, we leverage GPT-3.5 \citep{openai2023gpt4} to generate a set of semantically equivalent instructions $\mathcal{I}=\{I_1, ..., I_M\}$ by paraphrasing from an initial instruction $I_1$. In Fig.~\ref{fig:sensitivity}, we observe a severe fluctuation in human agreement scores by prompting Llama-3 8B~\citep{touvron2023llama} model with $C_{I_m\in\mathcal{I}}(x, I_m)$. This reflects a high prompt sensitivity and poor robustness of standard LLM evaluators. The observation aligns with previous research in position biases \citep{zhao2021calibrate}, and LLMs are sensitive to orders and formats of provided exemplars \citep{lu-etal-2022-fantastically, sclar2024quantifying}.

\rparagraph{Preference Fairness and Human Alignment}
\label{sec:distribution}
Following the previous finding, we hypothesize that the prompt sensitivity is mainly due to the preference bias incurred by spurious correlations from the instructions $\mathcal{I}$. We proceed to visualize the human agreement regarding preference distribution $p_I$ by different instructions $I$ across the entire query set $\{x_1, ..., x_N\}$, measured by $
p_{I, A} = \frac{1}{N} \sum_{i=1}^{N} \mathbb{I}\left(p(y_i = A|x_i, I) > p(y_i = B|x_i, I)\right)
$, where $\mathbb{I}(\cdot)$ is an indicator function that counts the number of predictions that candidate \textit{A} is preferred to \textit{B} in pairwise evaluations. In Fig.~\ref{fig:sensitivity}, we show that the patterns are nearly perfectly fitted to a quadratic regression function, where the highest human agreement point is close to $p_I = 0.5$, and instructions with more skewed decision distributions always degrade the evaluation alignment. Therefore, $p_I$ is a good indicator that connects decision fairness with human judgments, and instructions with \textit{fairer} decision preferences can lead to \textit{better} human-aligned LLM judgments.

\section{\ours: Zero-Shot Prompt Optimization with Fairer Preferences}
\sparagraph{Zero-Shot Fairness Learning}
Motivated by these findings, we now propose to automatically optimize the evaluation prompts for LLM evaluators toward fairer preferences, thereby achieving better human alignments. Importantly, the source preference distribution for an unbiased pairwise evaluator should naturally be uniform $p_S=1/{|\mathcal{Y}|}$ (by the law of large numbers) given a sufficient number of randomly sampled pairwise candidates. Consequently, we propose a zero-shot fairness learning objective function as $\texttt{fair}_{x_i\sim \mathcal{D}}(I) = -\frac{1}{J}\sum_{j=1}^{J} |p_S - p_{I, y_j}|$ in an unsupervised set of data $\mathcal{D}$ by measuring the absolute difference between the source prior and preference distribution. 

\rparagraph{Automatic Prompt Optimization}
\begin{algorithm}[t]
\label{alg:zepo}
\begin{footnotesize}
	    \caption{\ours.
	    }
	    \label{alg:main_alg}
	\begin{algorithmic}[1]
		\STATE \textbf{Input}: Initial instruction prompt $I$; LLM optimizer $\mathcal{O}$; LLM evaluator $\mathcal{E}$; unlabeled data $\mathcal{D}$; number of classes $J$; number of epochs $E$; population size $S$. 
		\STATE \textbf{Output}: Optimized Instruction prompt $I^*$
            \STATE {Initialize the instruction $I^*\leftarrow I$.}
		\FOR{ {$e$ in $E$} }
		\STATE {Obtain new instruction candidates from the LLM optimizer $\mathcal{O}$: $\mathcal{I}\leftarrow\mathcal{O}(I^*)$, where $|\mathcal{I}|=S$.}
                \FOR{ {$I\in\mathcal{I}$} }
                \STATE {LLM evaluator $\mathcal{E}$ generates a preference distribution over $\mathcal{D}$ (i.e., the decision rate for each class $y_i$), $p_{I, y_i}=\mathcal{E}(I)$, measured by the equation in Sec.~\ref{sec:distribution}.}
                \STATE {Compute the zero-shot fairness for each instruction candidate: $\texttt{fair}_{\mathcal{D}}(I) = -\frac{1}{J}\sum_{j=1}^{J}|\frac{1}{J} - p_{I, y_j}|$.}
                \ENDFOR
            \STATE {Update the best instruction: \\$I^{*} \leftarrow \operatorname*{arg\,max}_{I\in \mathcal{I}} \operatorname*{\texttt{fair}}_{\mathcal{D}}(I)$.}
		\ENDFOR
            \STATE {Return the optimized instruction $I^*$.}
	\end{algorithmic}
\end{footnotesize}
\end{algorithm}
In contrast with previous prompt optimization methods that heavily rely on labeled data, we propose \ours, an automatic \oursfullx framework. It is a more natural setup for reference-free LLM evaluations where human scores are usually unavailable in advance. 
\ours optimizes the evaluation prompts by maximizing the zero-shot fairness metric, such that $I^{*} = \operatorname*{arg\,max}_{I\in \mathcal{I}} \operatorname*{\texttt{fair}}_{x_i\sim \mathcal{D}}(I)$. We integrate an LLM paraphraser with a greedy search algorithm to update the instruction $I$ iteratively, where the detailed \ours algorithm is shown in Algorithm~\ref{alg:main_alg}. We refer to Appendix~\S\ref{sec:appendix} for more details on implementing \ours. It is worth noting that debiasing and calibration \citep{zheng2024large, zhou2024batch} methods can also control LLM evaluators for fairer preferences. We show in Figure~\ref{fig:permutate} that \ours is a meta-method orthogonal to existing debiasing approaches and leads to further improvements. In addition, we report the initial (seed) prompt and \ours-optimized prompt with corresponding fairness scores in Table~\ref{tab: prompts1} and \ref{tab: prompts2}.

\begin{table*}[t!]
\vspace{-1mm}
\centering
\def\arraystretch{1}
\setlength{\extrarowheight}{2pt}
\resizebox{\linewidth}{!}{%
\begin{tabular}{llllllllllll}
\hlinewd{1pt}
\multicolumn{1}{l}{\multirow{2}{*}{Models}} &   \multicolumn{4}{c}{News Room} &  \multicolumn{4}{c}{SummEval} & \multicolumn{1}{l}{\multirow{2}{*}{Avg.}}\\[1pt]
\multicolumn{1}{l}{}                        &  COH    & REL    & INF   & FLU   &   COH    & FLU   & CON   & REL &  \\
\cmidrule(lr){1-1} \cmidrule(lr){2-5} \cmidrule(lr){6-9} \cmidrule(lr){10-10}
\rowcolor{gray!20}
Other Metrics                                  &         &        &       &      &  &       &       &          &  \\
BertScore                                   & 0.15 & 0.16 & 0.13 & 0.17  & 0.28 &0.19& 0.11 & 0.31  &0.19\\
GPTScore                              & 0.31 & 0.35 & 0.26 & 0.31  & 0.28 & 0.31 & 0.38 & 0.22 &0.30\\
\cmidrule(lr){1-1} \cmidrule(lr){2-5} \cmidrule(lr){6-9} \cmidrule(lr){10-10}
\rowcolor{gray!20}
Mistral 7B                                 &       &               &      &  &       &       &       &    &  \\[-2pt]
\rowcolor{wan-brown}Scoring                                         & 0.32 & 0.39& 0.20  & 0.26  & 0.23  &0.19 & 0.37 &0.19&0.27\\
\rowcolor{wan-brown}G-Eval                                        & 0.36  & 0.36 &0.24 &  0.39   &  0.25 & \textbf{0.20}  & \textbf{0.39}   & 0.25&0.31\\
\rowcolor{wan-purple}Pairwise                                         & 0.33 &\textbf{0.40} &0.19 &0.19 &   0.06 & 0.01  & 0.07  & 0.16& 0.18\\
\rowcolor{wan-purple}\ours                                        & \textbf{0.47}\textcolor{racing-green}{\textit{+14\%}} & 0.38\textcolor{awesome-red}{\textit{-2\%}} & \textbf{0.44}\textcolor{racing-green}{\textit{+25\%}}& \textbf{0.48}\textcolor{racing-green}{\textit{+29\%}}  & \textbf{0.29}\textcolor{racing-green}{\textit{+23\%}} & 0.13\textcolor{racing-green}{\textit{+12\%}}&  0.32\textcolor{racing-green}{\textit{+25\%}}& \textbf{0.30}\textcolor{racing-green}{\textit{+14\%}}&\textbf{0.35}\textcolor{racing-green}{\textit{+17\%}}\\ \cmidrule(lr){1-1} \cmidrule(lr){2-5} \cmidrule(lr){6-9} \cmidrule(lr){10-10}
\rowcolor{gray!20}
Llama-3 8B                                 &       &        &             &  &       &       &       &     &  \\[-2pt]
\rowcolor{wan-brown}Scoring                                        & 0.42 &0.41&0.30 &0.29   & 0.35 & 0.23 &\textbf{0.32} & \textbf{0.46}  &0.35\\
\rowcolor{wan-brown}G-Eval                                       & 0.38 &0.34&0.26 &0.26  & 0.34 & 0.22 &0.29& 0.42 & 0.33\\ 
\rowcolor{wan-purple}Pairwise                                          & 0.49 & 0.51& 0.46&0.45  &0.24  & 0.12  & 0.30  & 0.21 &0.35\\
\rowcolor{wan-purple}\ours                                        & \textbf{0.57}\textcolor{racing-green}{\textit{+8\%}} & \textbf{0.54}\textcolor{racing-green}{\textit{+3\%}} & \textbf{0.55}\textcolor{racing-green}{\textit{+9\%}}& \textbf{0.56}\textcolor{racing-green}{\textit{+11\%}}&\textbf{0.40}\textcolor{racing-green}{\textit{+16\%}}  & \textbf{0.25}\textcolor{racing-green}{\textit{+13\%}} & 0.30\textcolor{racing-green}{\textit{+0\%}} & 0.39\textcolor{racing-green}{\textit{+18\%}} & \textbf{0.45}\textcolor{racing-green}{\textit{+10\%}}\\ \hlinewd{1pt}
\end{tabular}
}
\vspace{-2mm}
\caption{Spearman correlations on Mistral 7B and Llama-3 8B. We evaluate \bluebg{preference-based evaluators} and \yellowbg{direct-scoring evaluators} in terms of Coherence (COH), Relevancy (REL), Informativeness (INF), Fluency (FLU), and Consistency (CON). We highlight the \% improvement/degradation of \ours over ``\texttt{Pairwise}'' in \textcolor{racing-green}{+green}/\textcolor{awesome-red}{-red}. 
}
\label{tab:main}
\end{table*}

\section{Experiments and Results}
\sparagraph{Datasets and Models}
Following \citet{zhong-etal-2022-towards} and \citet{fu2023gptscore}, we evaluate \ours on representative meta-evaluation benchmarks, including two summarization tasks: News Room \citep{grusky-etal-2018-newsroom}
 and SummEval \citep{fabbri-etal-2021-summeval}, 
 and one dialog task: TopicalChat \citep{mehri-eskenazi-2020-usr} (see Appendix~\S\ref{sec:appendix} for further details). We examine \ours with state-of-the-art open-source LLMs, Mistral 7B~\citep{jiang2023mistral}
and Llama-3 8B~\citep{touvron2023llama}.

\rparagraph{Baselines}
We provide baseline scores for 
reference-free evaluators in the zero-shot setup, including \texttt{BERTScore}~\citep{Zhang2020BERTScore}, \texttt{GPTScore}~\citep{fu2023gptscore}, and \texttt{G-Eval}~\citep{liu-etal-2023-g}. \ours is applicable to state-of-the-art pairwise ranking evaluators, and we report experimental results from \texttt{Pairwise}~\citep{liu2024aligning} as the main baseline and provide direct scoring evaluation results named \texttt{Scoring} and \texttt{G-Eval} for reference.

\begin{figure}[!t]
    \centering
    \includegraphics[width=0.95\linewidth]{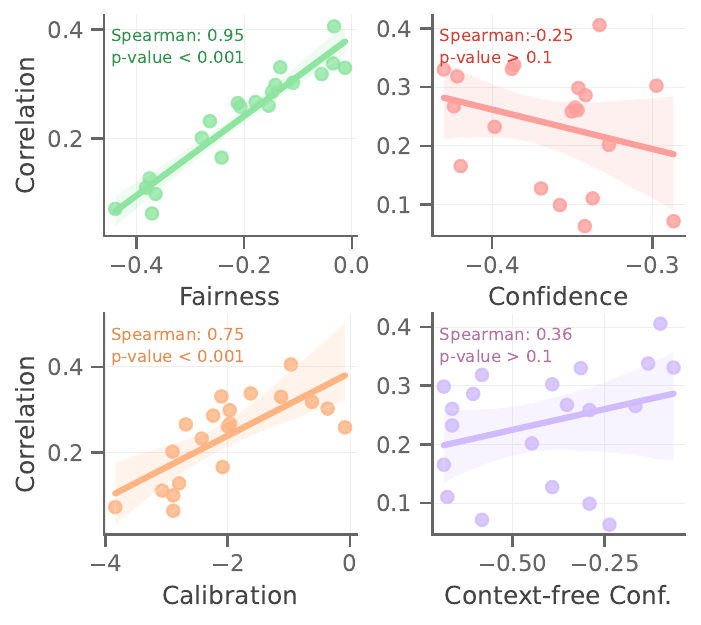}
    \vspace{-2.5mm}
    \caption{\textit{Fairness shows the strongest correlation with LLM evaluation performance.} Correlation studies of zero-shot learning objectives and LLM evaluation performance. The growth of the x-axis indicates better/stronger \textcolor{g-green}{fairness}, \textcolor{g-red}{confidence} (\textcolor{g-purple}{conf.}), and \textcolor{g-brown}{calibration}.}
    \label{fig:metrics}
    \vspace{-2mm}
\end{figure}
\rparagraph{Main Results}
We present \ours on representative meta-evaluation benchmarks in Table~\ref{tab:main}. Notably, \ours yields substantial gains in alignment with human judgments over almost all aspects on the \texttt{Pairwise} baseline: 17\% and 10\% on average on Mistral 7B and Llama-3 8B, respectively. It shows that manually designed evaluation criteria and instructions (without prompt optimization) can expose strong preference bias with LLM evaluators. By conducting \ours on \texttt{Pairwise} in a zero-shot setup, the performance of pairwise evaluators can be largely recovered, outperforming fine-calibrated direct scoring and the \texttt{G-Eval} baselines. Furthermore, we notice that weaker models, e.g. Mistral 7B, can exhibit more catastrophic evaluations, suffering from preference biases (e.g., on COH and CON aspects in SummEval), whereas Llama-3 8B generates relatively more robust evaluations. In both cases, \ours constantly mitigates the preference bias and better aligns LLM evaluators. Overall, the results indicate that \ours is a label-free and efficient prompt optimizer for effectively aligning LLM evaluators with human judgments.

\rparagraph{Zero-shot Learning Objectives}
We provide an in-depth analysis of the effectiveness of our proposed \texttt{Fairness} metric in comparison to other zero-shot objective functions as visualized in Fig.~\ref{fig:metrics}. We include model confidence, a commonly used zero-shot metric in exemplar selection \citep{lu-etal-2022-fantastically, wan-etal-2023-better, wan-etal-2023-universal}, measured as the negative of entropy. Calibration-based approaches have been effective in mitigating position biases \citep{zhao2021calibrate, wang2023large}. We adopt a zero-shot calibration metric from Batch Calibration \citep{zhou2024batch} and context-free confidence as another metric from Fair-Prompting~\citep{ma2024fairness}, where overconfidence is argued to result in unfairness. First, \texttt{Fairness} shows the largest Spearman correlation with LLM evaluation performance, guaranteeing its effectiveness with \ours. Following fairness, \texttt{Calibration} is more weakly correlated, whereas \texttt{Confidence} metrics fail to serve as good objectives for \ours, with poorer correlations. 
\begin{figure}[t]
    \centering
    \begin{subfigure}[b]{0.445\linewidth}
        \centering
        \includegraphics[width=\textwidth]{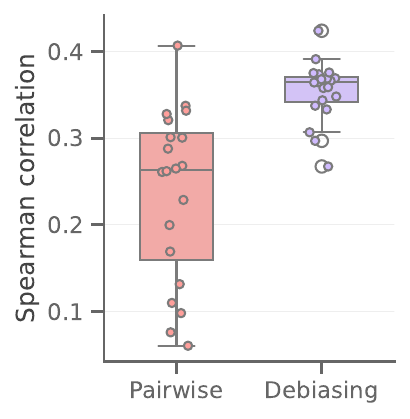}
        \caption*{}
    \end{subfigure}%
    \begin{subfigure}[b]{0.555\linewidth}
        \centering
        \includegraphics[width=\textwidth]{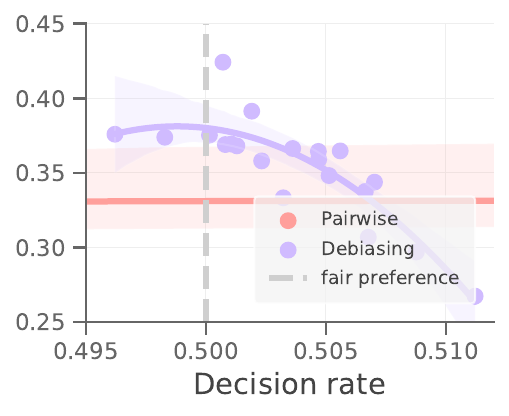}
        \caption*{}
    \end{subfigure}
    \vspace{-11mm}
    \caption{\textit{\ours is orthogonal to debiasing approaches and brings further improved LLM judgments.} Sensitivity and evaluation performance studies on preference fairness before and after applying permutation debiasing on the COH aspect in SummEval from Llama-3 8B.}
    \label{fig:permutate}
\end{figure}

\rparagraph{Complementarity with Debiasing}
We further extend our study of \ours, focusing on its orthogonality/complementarity with debiasing approaches. We implement the \textit{permutation debiasing} method which averages the probability for different orders/positions of the same candidates, also termed Balanced Position Calibration~\citep{wang2023large}. Fig.~\ref{fig:permutate} shows that the \texttt{Debias} method first improves the lower bar of the evaluation performance of LLMs. Secondly, when we inspect the preference distribution after applying \texttt{Debias}, we observe a fairer preference distribution where the decision rates become much closer to 0.5. However, LLM evaluators are still sensitive to semantically equivalent instructions even after debiasing, where the judgment alignment varies substantially from 0.26 to 0.43. In addition, we observe a similar quadratic curve in the second plot, indicating that our previous findings still hold: \textit{fairer preferences lead to improved human-aligned LLM judgments}. 

Following this observation, we conduct additional experiments on \ours \textit{with} and \textit{without} permutation debiasing. Table~\ref{tab:permutation} shows that further gains can be achieved by integrating debiasing methods with prompt optimization. Therefore, we conclude that \ours is a meta-method on zero-shot prompt optimization while being orthogonal to other debiasing and calibration methods. In light of this work, we expect to build toward improved human-aligned LLM evaluators with a combination of prompt optimization, calibration, and advanced debiasing methods.

\begin{table}[t]
\centering
\def\arraystretch{1.3}
\setlength{\extrarowheight}{2pt}
\setlength{\tabcolsep}{3pt}
\resizebox{\linewidth}{!}{%
\begin{tabular}{llllll}
\hlinewd{1pt}
\multicolumn{1}{l}{\multirow{2}{*}{Methods}} &   \multicolumn{4}{c}{News Room} &   \multicolumn{1}{l}{\multirow{2}{*}{Avg.}}\\[1pt]
\multicolumn{1}{l}{}                        &  COH    & REL    & INF & FLU    &  \\
\cmidrule(lr){1-1} \cmidrule(lr){2-5} \cmidrule(lr){6-6}

Pairwise                                  &   0.49    &        0.51       &  0.46    &0.45 &0.48\\
\ours                                  &   0.57    &        0.54       &  0.55    &0.56 &0.56\\ \hline
Pairwise + Debias                                  &   0.60    &        \textbf{0.61}       &  0.64    &\textbf{0.58} &0.61\\
\ours + Debias                                 &   \textbf{0.64}  \textcolor{racing-green}{\textit{+4\%}}  &        \textbf{0.61} \textcolor{racing-green}{\textit{+0\%}}      &  \textbf{0.72} \textcolor{racing-green}{\textit{+8\%}}   &0.57\textcolor{awesome-red}{\textit{-1\%}} &\textbf{0.64}\textcolor{racing-green}{\textit{+3\%}}\\ 

\hlinewd{1pt}
\end{tabular}
}
\caption{Spearman correlations on News Room with Llama-3 8B before and after applying permutation debiasing. We highlight the \% improvement/degradation of \ours over ``\texttt{Pairwise}'' after debiasing in \textcolor{racing-green}{+green}/\textcolor{awesome-red}{-red}. 
}
\label{tab:permutation}
\end{table}

\section{Conclusion}
We first analyzed the relationship between preference fairness and human alignment; it revealed that LLM evaluators produce highly skewed preference distributions even with semantically equivalent instructions. We further showed that fairer preferences can yield improved human-aligned LLM judgments. Based on this insight, we proposed a zero-shot prompt optimization framework with a fairness-aware zero-shot proxy. It substantially improves alignments of pairwise LLM evaluators with humans, without any labeled data, and serves as a meta-method orthogonal to debiasing approaches. 

\section*{Limitations}
First, \ours is a zero-shot method that learns the zero-shot fairness metric from unlabeled data. It still requires a sufficient number of random \textit{unlabeled} samples for pairwise evaluations to obtain a good estimation of preference distribution for fairness. We argue that such a data requirement is mild, as in the evaluation setup, the bottleneck lies in human-annotated labels, not unlabeled inputs. %
Second, \ours is primarily designed for preference-based evaluators, and we have widely examined the effectiveness of \ours in pairwise evaluations. Though pairwise evaluation appears to be the current leading standard, it is possible that future advances in LLM evaluators can achieve more efficient evaluation-by-ranking in multi-choice question formats with more than two classes, which have not been included in our current study. However, in principle, the proposed zero-shot fairness objective is a general learning metric scalable to any number of classes based on its uniform prior. 

Lastly, \ours only integrates a basic LLM optimizer in exploring instruction candidates at a paragraph level with a greedy search algorithm. However, \ours is a meta-framework also orthogonal to LLM optimizers with more advanced search algorithms, and this synergy warrants further investigation in future work. \ours serves as a first step towards LLM evaluation with fairer preferences and is easy to extend with more exploitation-driven LLM optimizers in alternative search spaces.

\section*{Acknowledgements}
The work has been supported by the UK Research and Innovation (UKRI) Frontier Research Grant EP/Y031350/1 (the UK government’s funding guarantee for ERC Advanced Grants) awarded to Anna Korhonen at the University of Cambridge. The work has also been supported in part by a Royal Society University Research Fellowship (no 221137; 2022-) awarded to Ivan Vuli\'{c}, and by the UK EPSRC grant EP/T02450X/1.
\bibliography{custom, anthology}
\appendix

\clearpage
\section{Implementation Details}
\label{sec:appendix}

\sparagraph{\ours}
In this section, we include implementation details to enable the reproducibility of our work. Regarding the template and prompt across all the experiments reported, we use the prompt template from Table~\ref{tab: templates}. ZEPO evaluation results are conducted on top of the state-of-the-art pairwise evaluator, PairS \citep{liu2024aligning}, which leverages pairwise comparisons between randomly sampled pairs and aggregates them into a ranked sequence with a sorting-based search algorithm. We use GPT-3.5-turbo as the LLM optimizer with a temperature of 0.9, which is instructed to generate diverse and creative paraphrasing of the initial instruction. Following that, we implement Mistral-7B-Instruct-v0.1 and Meta-Llama-3-8B-Instruct as our main LLM evaluators. In practice, we set 5 epochs with a population size $S$ of 5 that sufficiently converges to the fairest instruction. For $|\mathcal{D}|$, we use 2,400 pairwise sampling (10 data points) per instruction for SummEval, 840 (20 data points) for News Room, and 1,200 (60 data points) for TopicalChat based on the number of candidates per data point. \ours serves as a first step towards fairer LLM evaluations, and we defer
investigations on \ours with tighter, more sampling-efficient constraints to future work. 
\begin{table}[t]
\centering
\def\arraystretch{1.2}
\setlength{\extrarowheight}{2pt}
\setlength{\tabcolsep}{3pt}
\resizebox{\linewidth}{!}{%
\begin{tabular}{lllll}
\hlinewd{1pt}
\multicolumn{1}{l}{\multirow{2}{*}{Models}} &   \multicolumn{3}{c}{TopicalChat} &   \multicolumn{1}{l}{\multirow{2}{*}{Avg.}}\\[1pt]
\multicolumn{1}{l}{}                        &  NAT    & ENG    & OVE    &  \\
\cmidrule(lr){1-1} \cmidrule(lr){2-4} \cmidrule(lr){5-5}

\rowcolor{gray!20}
Mistral 7B                                 &       &               &      &  \\[-2pt]
Pairwise                                  &   0.13    &        0.18       &  0.22    &0.18\\
\ours                                 &   \textbf{0.14}  \textcolor{racing-green}{\textit{+1\%}}  &        \textbf{0.25} \textcolor{racing-green}{\textit{+7\%}}      &  \textbf{0.28} \textcolor{racing-green}{\textit{+6\%}}   &\textbf{0.23}\textcolor{racing-green}{\textit{+5\%}}\\ \cmidrule(lr){1-1} \cmidrule(lr){2-4} \cmidrule(lr){5-5}

\rowcolor{gray!20}
Llama-3 8B                                 &       &        &             &\\[-2pt]
Pairwise                                     &   0.02    &     0.08          &  0.14    &0.05\\
\ours                                        &    \textbf{0.16} \textcolor{racing-green}{\textit{+14\%}}  &     \textbf{0.26} \textcolor{racing-green}{\textit{+18\%}}         &   \textbf{0.46} \textcolor{racing-green}{\textit{+32\%}}  &\textbf{0.30}\textcolor{racing-green}{\textit{+25\%}}\\ \hlinewd{1pt}
\end{tabular}
}
\vspace{-1mm}
\caption{Spearman correlations on TopicalChat with Mistral and Llama-3. We evaluate in terms of Naturalness (NAT), Engagement (ENG), and Overall quality (OVE). We highlight the \% improvement/degradation of \ours over ``Pairwise'' in \textcolor{racing-green}{+green}/\textcolor{awesome-red}{-red}. 
}
\label{tab:topical}
\end{table}

\rparagraph{Zero-Shot Learning Objectives}
Entropy is a commonly used zero-shot metric: $-\sum_{j}p_j\operatorname*{log}p_j$. In Fig.~\ref{fig:metrics}, we use entropy as a confidence measurement for LLM evaluators and treat $\texttt{Confidence} = \sum_{j}p_j\operatorname*{log}p_j$ in the negative of entropy averaged across $\mathcal{D}$. However, in the context of LLM evaluations, overconfidence may further misalign LLM evaluators with human judgments. Context-free confidence is computed with the same formulation above but with a content-free input $C_I(\text{[N/A]}, I)$ adopted from the contextual calibration \citep{zhao2021calibrate}. Context-free confidence is introduced in Fair-Prompting~\citep{ma2024fairness}, where the main idea is to select exemplars with the lowest confidence with respect to a content-free input, such that the prediction for classes is more balanced with the prompt template alone. In addition, we adopted a zero-shot calibration metric from Batch Calibration \citep{zhou2024batch}: $\texttt{Calibration}= -|\frac{1}{N}\sum( \operatorname{log}p_A - \operatorname{log}p_B)|,$ which measures the absolute distance in the marginalized logits between two classes. 
\lstset{basicstyle=\footnotesize}
\begin{table}[t]
\centering
\setlength{\extrarowheight}{2pt}
\renewcommand{\arraystretch}{1}
\setlength{\tabcolsep}{5pt}
\begin{tabular}{p{\linewidth}}
\hlinewd{1pt}
Prompt Templates for \texttt{Pairwise} and \ours in summarization.   \\\hline
\begin{lstlisting}             
Source text: [SOURCE_TEXT]

Summary A: [SUMMARY_1]

Summary B: [SUMMARY_2]

Question: [INSTRUCTION]
Answer: [OUTPUT]
\end{lstlisting}
\\ \hline
Prompt templates for \texttt{Pairwise} and \ours in dialog.   \\\hline
\begin{lstlisting}                   
Dialog history: [DIALOG_HISTORY]

Response Candidate A: [RESPONSE_1]

Response Candidate B: [RESPONSE_2]

Question: [INSTRUCTION]
Answer: [OUTPUT]
\end{lstlisting}
\\ \hline
Prompt templates for LLM Optimizer to generate new instruction candidates.   \\\hline
\begin{lstlisting}                   
Paraphrase the following instruction 
for a pairwise comparison task. 
Do not change the keyword "[ASPECT]". 
Be diverse and creative in paraphrasing. 
Return the instruction only. 

Input: [INSTRUCTION]

Output: [NEW_INSTRUCTION]
\end{lstlisting}
\\ \hlinewd{1pt}
\end{tabular}

\caption{Prompt template for pairwise comparisons and the LLM optimizer to generate paraphrased instructions.}
\label{tab: templates}

\end{table}

It indicates a uniform prior in the logit space, and a better-calibrated model can generate fairer predictions in terms of their scores. In contrast with calibration, our fairness metric is based on a uniform prior in the preference (decision) distribution and demonstrates the strongest correlation with LLM evaluation performance.

\rparagraph{Pointwise Baselines} We implement two pointwise evaluator baselines: direct \texttt{Scoring} and \texttt{G-Eval}. For both cases, the LLM evaluators are tasked with rating a specific aspect of the output candidate using an integer score on the Likert scale \citep{likert1932technique}. In the \texttt{Scoring} approach, the evaluators assign a single score with the highest predictive probability to each output candidate. For the \texttt{G-Eval} baseline, the final score is calculated by taking the weighted average of the scores across all five score tokens. We use the same prompt templates and evaluation criteria from previous work \citep{liu-etal-2024-calibrating-llm}, which have been calibrated and deliver robust evaluations. As indicated in the main paper, \ours shows improved evaluation results in general over the aforementioned calibrated baselines.

\begin{table*}[h!]
\centering
\begin{tblr}{
  width = \linewidth,
  colspec = {Q[83]Q[688]Q[220]},
  hline{1-11} = {-}{},
  hline{3-10} = {-}{0.03em},
  hline{1,5} = {-}{0.08em},
}
Aspect          & Instruction Prompt                                                                                & Fairness           \\
COH            & {\textcolor{g-red}{Initial Prompt}: Evaluate and compare the coherence of the two summary candidates for the given source text. 
                        Consider coherence aspects such as clarity and logical flow. 
                        A summary is coherent if it accurately captures the key information from the article, and presents them in a clear manner. 
                        Which summary candidate has better coherence? 
                        If the candidate A is better, please return 'A'. 
                        If the candidate B is better, please return 'B'. 
                        You must return the choice only. \\
                        
                       \textcolor{g-green}{\ours-Optimized Prompt}: Assess and contrast the coherence of the two summaries using the provided text. Take into account clarity and logical progression. A coherent summary efficiently conveys the main details from the text in a clear and organized manner. Which summary demonstrates stronger coherence? Select 'A' for option A or 'B' for option B. Indicate your chosen option.
 }                                           & {Initial:
 -0.288   \\
 \\
 Optimized:
 \textbf{-0.007}} \\
FLU & {\textcolor{g-red}{Initial Prompt}: Evaluate and compare the fluency of the two summary candidates for the given source text. \
                        Which summary candidate has better fluency? \
                        If the candidate A is better, please return 'A'. \
                        If the candidate B is better, please return 'B'. \
                        You must return the choice only.\\
                        
                        \textcolor{g-green}{\ours-Optimized Prompt}: Evaluate the smoothness of each summary choice using the given text. Decide which summary showcases better fluency. Choose 'A' for candidate A or 'B' for candidate B. Please only submit your chosen option.
}& {Initial:
 -0.417   \\
 \\
 Optimized:
 \textbf{-0.018}}    \\
CON             & {\textcolor{g-red}{Initial Prompt}: Evaluate and compare the consistency of the two summary candidates for the given source text. \
                        A summary is consistent with the article if it faithfully reflects the main points, facts, and tone of the article. \
                        A summary is inconsistent if it introduces any errors, contradictions, or distortions of the original article. \
                        Which summary candidate has better consistency? \
                        If the candidate A is better, please return 'A'. \
                        If the candidate B is better, please return 'B'. \
                        You must return the choice only.\\
                        
                        \textcolor{g-green}{\ours-Optimized Prompt}: Evaluate the consistency of two different ways of summarizing the given text. Find the summary that best captures the main ideas, details, and tone of the original text. Note any mistakes or differences in the summaries. Choose either 'A' for option A or 'B' for option B as the superior choice. Share your selected option.
}& {Initial:
 -0.295   \\
 \\
 Optimized:
 \textbf{-0.012}}    \\

\end{tblr}
\caption{Initial prompt and the \ours-found prompt. We report the fairness metric before and after optimization.}
\label{tab: prompts1}

\end{table*}

\begin{table*}[!h]
\centering
\begin{tblr}{
  width = \linewidth,
  colspec = {Q[83]Q[688]Q[220]},
  hline{1-11} = {-}{},
  hline{3-10} = {-}{0.03em},
  hline{1,4} = {-}{0.08em},
}
Aspect          & Instruction Prompt                                                                                & Fairness           \\
REL            & {\textcolor{g-red}{Initial Prompt}: Evaluate and compare the relevance of the two summary candidates for the given source text. \
                        A summary is relevant if it captures the main points from the article, without leaving out any crucial details or adding any unnecessary or inaccurate ones. \
                        A summary is more relevant if it uses the same or similar terms and expressions as the article. \
                        A summary is less relevant if it omits some of the key facts from the article, or if it introduces irrelevant information that is not supported by the article. \
                        Which summary candidate has better relevance? \
                        If the candidate A is better, please return 'A'. \
                        If the candidate B is better, please return 'B'. \
                        You must return the choice only.\\
                        
                       \textcolor{g-green}{\ours-Optimized Prompt}: Assess the relevance of the two summaries presented for the text and pick the one that closely matches the main points of the article using similar language. Select 'A' for candidate A or 'B' for candidate B. Display your selection.
 }                                           & {Initial:
 -0.3625   \\
 \\
 Optimized:
 \textbf{-0.0003}} \\
INF & {\textcolor{g-red}{Initial Prompt}: Evaluate and compare the informativeness of the two summary candidates for the given source text. \
                        Evaluate how each summary converts their input text to natural language text, without omitting, adding, or distorting any facts. \
                        Which summary candidate has better informativeness? \
                        If the candidate A is better, please return 'A'. \
                        If the candidate B is better, please return 'B'. \
                        You must return the choice only.\\
                        
                        \textcolor{g-green}{\ours-Optimized Prompt}: Assess and contrast the informativeness of two summaries based on the provided source material. Examine how accurately each summary reflects the original content. Determine which summary is more informative by selecting either 'A' or 'B'. Only indicate your choice.
}& {Initial:
 -0.217   \\
 \\
 Optimized:
 \textbf{-0.001}}    \\

\end{tblr}
\caption{Initial prompt and the \ours-found prompt. We report the fairness metric before and after optimization.}
\label{tab: prompts2}

\end{table*}

\end{document}